\documentclass[conference]{IEEEtran}
\IEEEoverridecommandlockouts
\usepackage{makecell}
\usepackage{adjustbox}
\usepackage{cite}
\usepackage{amsmath,amssymb,amsfonts}
\usepackage{algorithmic}
\usepackage{graphicx}
\usepackage{textcomp}
\usepackage{xcolor}
\usepackage{booktabs}
\usepackage{array}
\usepackage{url}
\usepackage{multirow} 
\usepackage[colorlinks=true, linkcolor=blue, citecolor=blue, urlcolor=blue]{hyperref}
\usepackage{float}
\usepackage{caption}
\usepackage{subcaption}
\usepackage{tikz}
\usetikzlibrary{shapes, arrows, positioning, fit}
\def\BibTeX{{\rm B\kern-.05em{\sc i\kern-.025em b}\kern-.08em
		T\kern-.1667em\lower.7ex\hbox{E}\kern-.125emX}}

\begin{document}
	
	\title{Beyond Similarity: Heterogeneous Graph Learning for Multi-Objective Food Substitution in Charitable Food Agencies}
	\author{
    \IEEEauthorblockN{Naimur Rahman Chowdhury}
    \IEEEauthorblockA{
    \text{The work was done prior to joining Amazon.} \\
    \textit{Department of Industrial and Systems Engineering} \\
    \textit{North Carolina State University} \\
    Raleigh, NC, USA \\
    nchowdh2@ncsu.edu
    }
    \and
    \IEEEauthorblockN{Limon Bin Hossain}
    \IEEEauthorblockA{
    \textit{Department of Industrial and Production Engineering} \\
    \textit{Bangladesh University of Engineering and Technology} \\
    Dhaka, Bangladesh \\
    mdlimonbinhossain@gmail.com
    }
    }
	
	\maketitle
	\begin{abstract}
		Charitable food agencies play an important role in alleviating food insecurity by distributing donated food to people in need. However, they rely on ad hoc in-kind donations and often face shortages of specific foods, so they offer substitutes. A good food substitution requires matching household preferences, nutritional needs, and item similarity. Agencies have limited direct records of consumption behavior due to resource constraints, making it challenging to make an appropriate substitution decision that meets multiple criteria. In this study, we propose a heterogeneous graph neural network (HeteroGNN), a source-grounded recommendation framework for food substitution in charitable food agencies. We first build a unified relational graph from large-scale public data sources, combining household behavior on food consumption and food nutrient information in the United States (US) context. We treat the substitution recommendation as a multi-objective ranking problem with three targets, including behavior affinity, health suitability, and substitution similarity. We train and validate the proposed framework under standard graph relationship and adverse cold-start settings by removing relational edges from the graph. Our results show that the proposed framework leverages relational information beyond node features in predicting consumption behavior. Additionally, the proposed framework remains robust with sparsity when the model receives incomplete information about behavior and nutrient features. Finally, we show the weak correlation among different objectives, thereby justifying the multi-objective framing as a replacement for an aggregated decision. The proposed framework can help downstream charitable agency decision-makers make context-specific substitution recommendations with limited information available.  
	\end{abstract}
	
	
\section{Introduction}
In the United States (US), one in six people is served by food banks
and their associated partner agencies, such as food pantries and soup kitchens, which play an important role in reducing food insecurity \cite{park2024adverse}. However, these charitable food agencies are heavily dependent on external in-kind donations, which limit their ability to cater to food-insecure communities \cite{chowdhury2025influence, sharma2021data}. Hence, these organizations often have to provide substitutes for requested or preferred items to food-insecure households from their limited catalog \cite{sengul2023equitable, bazerghi2016role}. However, being a good food substitute requires several components, including familiarity within the household, nutritional suitability for household members, and similarity to the requested item \cite{short2022food,simmet2017nutritional}. This substitution decision for a charitable agency volunteer is non-trivial, since households
served are heterogeneous in demographic characteristics and often have little or no directly observed preference history to the agencies with limited data management capacity \cite{hasnain2021elicitation}. Public data on household food consumption provide dietary patterns and familiarity \cite{cdc2024nhanes,usda2026foodaps}, but no current public dataset provides structured records of household context, requested foods, and offered substitutes, so a recommendation system may rely on indirect behavioral signals.

The most relevant studies on charitable food agencies have focused on forecasting
donation quantity and timing \cite{davis2016analysis,sharma2021data,sharma2025foodrl}, but modeling the substitution decision itself still remains unexplored. In food substitution recommendations for requested items, to match preferences across multiple dimensions (affinity, nutritional values, and similarity), the system needs to rely on behavioral evidence, keeping familiarity, health, and similarity rather than a general recommender that is trained on direct user-item interaction \cite{kipf2017semi,hamilton2017inductive} to optimize a single score \cite{ricci2015recommender,burke2002hybrid}. This process generalizes to foods and household groups that it has not directly observed.

To this end, although the final substitution score can be analytically obtained when all relevant information is available, the challenge arises 
with sparse, highly interdependent inputs that are scattered across separate data sources. Moreover, an analytical form struggles to produce a recommendation score when the input is incomplete. In this study, we propose a heterogeneous graph neural network (HeteroGNN) to complement heuristic alternatives requiring enumeration in every combination of nutrition, consumption behavior, and food similarity that
becomes harder to maintain as the number of foods, household groups, and
relations grow. We leverage extensive public data sources \cite{usda2026fdc,usda2026foodaps,cdc2024nhanes} to extract household consumption behavior, nutritional information, and similarity. Finally, we propose a recommendation framework that produces a ranked list of substitutes with three objectives, i.e., behavior affinity, health suitability, and
substitution similarity. The objectives of this study are as follows.

    \begin{itemize}
        \item To develop a heterogeneous graph framework that integrates a sparse public dataset into a unified relational structure for food substitute recommendation.
        \item  To jointly learn behavior affinity, health suitability, and substitution similarity as complementary objectives, preserving downstream decision makers' ability to evaluate trade-offs in recommendations.
        \item To evaluate the robustness of the learning framework in reliable recommendation performance under incomplete and sparse information, where traditional systems struggle.
    \end{itemize}

	\section{Methods}
	\label{sec:methods}
	We formulate a decision problem where $\mathcal{U}$ denotes the set of household-group contexts, and $\mathcal{F}$ the set of foods in the agency catalog. Here, $\mathcal{R}\subseteq\mathcal{F}$ is the set of possible requested foods by a household. For a request from context $u\in\mathcal{U}$ for food $r\in\mathcal{R}$, the system receives a candidate set
	$\mathcal{C}(r)\subseteq\mathcal{F}$ of substitutes. We aim to
	rank the candidates $f\in\mathcal{C}(r)$ along three objectives shown in Eq. \ref{eq1}.
	\begin{equation} \label{eq1}
		S(u,r,f)=\bigl(S_b(u,f),\,S_h(u,f),\,S_s(r,f)\bigr),
	\end{equation}
	where $S_b$ is behavior affinity, $S_h$ is health suitability, and $S_s$ is
	substitution similarity. Our learning model estimates $\hat{S}_b$, $\hat{S}_h$, and
	$\hat{S}_s$ separately.

\subsection{Dataset and Graph Construction}
\begin{figure}[t]
\centering
\includegraphics[width=\columnwidth]{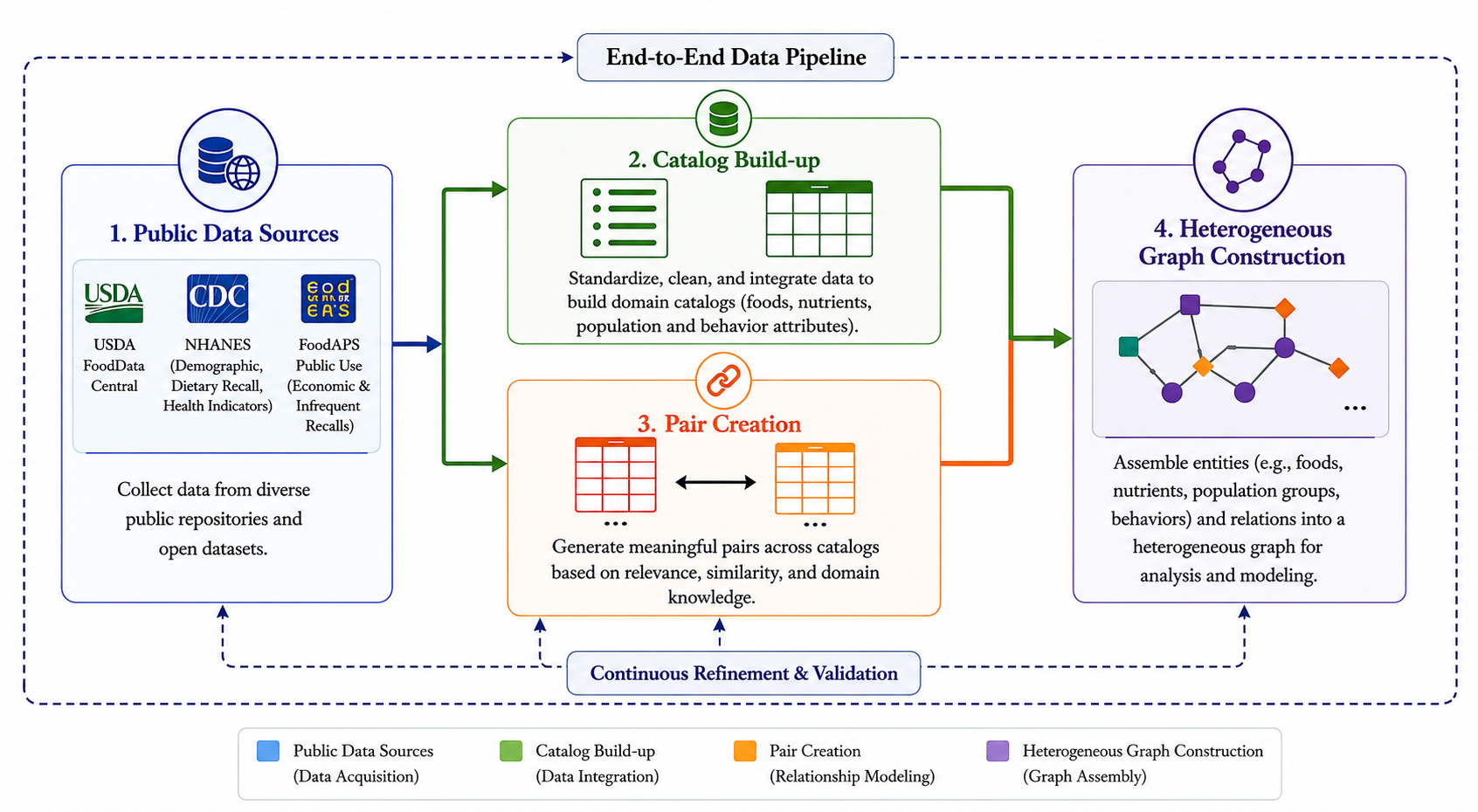}
\caption{Complete data construction with food catalog, population, and behavior relations for a heterogeneous graph.}
\label{fig:pipeline}
\end{figure}

We construct the heterogeneous graph from four public datasets through two streams, as shown in Figure \ref{fig:pipeline}. The first stream builds the food catalog and substitution pairs. Starting with generic food items relevant to charitable food agencies \cite{levi2022nutrition}, we map each item to its corresponding USDA FoodData Central Foundation Foods record \cite{usda2026fdc} for nutrient profiles, supplementing missing nutrients with the Food and Nutrient Database for Dietary Studies (FNDDS) \cite{usda2024fndds}. We construct directional substitution pairs based on culinary-use overlap, food category, form, and nutrient closeness. The second stream builds population and behavioral relations using the National Health and Nutrition Examination Survey (NHANES) \cite{cdc2024nhanes}, which provides demographic, dietary, and health indicators. Moreover, FoodAPS \cite{usda2026foodaps} provides records of food acquisition. We group households by income band, household size, and SNAP participation, aggregating NHANES consumption and FoodAPS acquisition records into group-level behavioral evidence. Additionally, we incorporate nutrition guidance from Healthy Eating Research \cite{her2020guidelines} to rank foods for charitable food supplies.

The constructed heterogeneous graph has four node types, i.e., household groups, foods, nutrients, and categories. Additionally, it constructs four relational families, i.e., group--food behavior, food--nutrient content, food--category membership, and food--food substitution similarity. The graph covers 26,250 people across 16,759 household/respondent contexts, aggregated into 42 household groups and 9,903 foods, with 102,441 food--nutrient records. Behavioral evidence from 188,148 consumption lines and 254,587 acquisition items results in 59,552 group-food behavior edges.

\subsection{Objective Targets}

In this study, we use proxy supervision targets based on observed consumption, nutritional composition, and food attributes, as direct records for charitable agencies are limited.

\subsubsection{Behavior Affinity Target}
For each observed household-group--food pair $(u,f)$, the behavior target is the log-scaled count of observed occasions, normalized within the group in Eq.~\ref{eq.behav}, which follows implicit feedback \cite{hu2008collaborative}.
\begin{equation} \label{eq.behav}
	y_b(u,f)=\frac{\log\bigl(1+\text{occasions}_{u,f}\bigr)}
	{\max_{f'}\log\bigl(1+\text{occasions}_{u,f'}\bigr)}.
\end{equation}
For the binary behavior head, observed pairs are positives and unobserved pairs are sampled as negatives, as in Bayesian Personalized Ranking (BPR) \cite{rendle2009bpr}.

\subsubsection{Health Suitability Target} \label{health_suit}
The relational database gives six profile-specific health scores, i.e., general, diabetes, hypertension, obesity, child nutrition, and senior nutrition, for members in a household group. Each food-profile score is a nutrient-density score based on beneficial and limiting nutrients, following practical guidance for charitable food environments \cite{her2020guidelines}. Beneficial nutrients are mapped as in Eq. \ref{ben_nut}, whereas Eq. \ref{lim_nut} maps limiting nutrients. 
\begin{equation} \label{ben_nut}
	c^{+}(f,n)=\min\!\left(\frac{\max(a_{f,n},0)}{\rho_n},\,1\right),
\end{equation}

\begin{equation}\label{lim_nut}
	c^{-}(f,n)=1-\min\!\left(\frac{\max(a_{f,n},0)}{\rho_n},\,1\right),
\end{equation}
where $a_{f,n}$ is the per-100\,g nutrient amount and $\rho_n$ is the profile reference amount. 

The profile-specific score is defined in Eq. \ref{prof_health} that follows the Nutrient-Rich Foods (NRF) index, which scores foods on a per-reference-amount basis \cite{drewnowski2009defining,fulgoni2009development,drewnowski2010nutrient}.
Here, $c(f,n)=c^{+}(f,n)$ for beneficial nutrients and $c(f,n)=c^{-}(f,n)$ for limiting nutrients.
\begin{equation} \label{prof_health}
	H(f,p)=\frac{\sum_{n\in\mathcal{N}_f}w_{p,n}\,c(f,n)}
	{\sum_{n\in\mathcal{N}_f}w_{p,n}}.
\end{equation}
For household group $u$, the profile mixture $\pi_{u,p}$ is estimated
from observed health indicators in the group. The final group--food health target is defined in Eq. \ref{fina_health}.
\begin{equation}\label{fina_health}
	y_h(u,f)=\sum_{p\in\mathcal{P}}\pi_{u,p}H(f,p).
\end{equation}

\subsubsection{Substitution Similarity Target}

From the food catalog, we construct 124 directional substitution pairs, of which 56 pairs are mapped to the heterogeneous graph containing 9,903 food nodes and used for similarity supervision. HeteroGNN learns to predict similarity scores, $\hat{S}_s(r,f)\in[0,1]$, for other candidate food pairs without explicit labels.

\subsection{HeteroGNN}
For the HeteroGNN, a type-specific encoder maps each node's features to $h^{(0)}_{v}=\sigma\bigl(W_{\tau(v)}x_{v}+b_{\tau(v)}\bigr)$. Each message-passing layer $\ell$ then aggregates neighbors with relation-specific weights as shown in Eq. \ref{eq:encod}.
\begin{equation}
 \label{eq:encod}
	\begin{split}
		h^{(\ell+1)}_{v}=\mathrm{ReLU}\Bigl(\mathrm{LN}\Bigl(&W^{\text{self}}_{\tau(v)}h^{(\ell)}_{v}\\
		&+\sum_{\rho\in\mathcal{R}_G}\ \sum_{u\in\mathcal{N}_{\rho}(v)}
		\omega^{\rho}_{u,v}W_{\rho}h^{(\ell)}_{u}\Bigr)\Bigr),
	\end{split}
\end{equation}
Here, $\rho$ indexes graph relations, $\omega^{\rho}_{u,v}$ is the edge weight, and LN is layer normalization. Stacking these layers gives an inductive node representation \cite{kipf2017semi,hamilton2017inductive}.

After encoding, three feed-forward heads with sigmoid outputs score node pairs from concatenated embeddings with $\hat{S}_b(u,f)=g_b([h_u,h_f])$, $\hat{S}_h(u,f)=g_h([h_u,h_f])$, and $\hat{S}_s(r,f)=g_s([h_r,h_f])$. 

Finally, the model minimizes $\mathcal{L}=\lambda_b\mathcal{L}_b+\lambda_h\mathcal{L}_h+\lambda_s\mathcal{L}_s$, where $\mathcal{L}_b$ is binary cross-entropy on behavior labels and $\mathcal{L}_h$ and $\mathcal{L}_s$ are mean-squared errors on the health and similarity targets.
	
	\subsection{Weighted Multi-Objective Score}
	The model produces three predicted scores per candidate. A weighted score is used for the composite scoring as in Eq. \ref{eq:weighted_score}.
	\begin{equation}
		R(u,r,f)=\alpha_b\hat{S}_b(u,f)+\alpha_h\hat{S}_h(u,f)
		+\alpha_s\hat{S}_s(r,f),
		\label{eq:weighted_score}
	\end{equation} 
	  $\alpha_b$, $\alpha_h$, and $\alpha_s$ are the weights for each objective, providing downstream decision-makers with the flexibility to evaluate trade-offs among recommendations. For this study, we use a uniform weight across the three targets.
\begin{table*}[!t]
\centering
\caption{Performance comparison under the \textbf{standard} evaluation setting.}
\label{tab:standard_results}
\begin{tabular}{llcccc}
\toprule
\multicolumn{2}{l}{\textbf{Metric}} &
\textbf{HeteroGNN} &
\textbf{MLP} &
\textbf{BPR} &
\textbf{LightGCN} \\
\midrule

\textit{Behavior affinity}
& ROC-AUC        & \textbf{0.9480} & 0.9009 & 0.9144 & 0.9369 \\
& PR-AUC         & \textbf{0.9027} & 0.7809 & 0.8467 & 0.8838 \\
& Precision@5    & \textbf{0.9952} & 0.8000 & 0.9667 & 0.9667 \\
& Recall@5       & \textbf{0.0262} & 0.0186 & 0.0244 & 0.0240 \\
& NDCG@5         & \textbf{0.9965} & 0.8039 & 0.9724 & 0.9699 \\

\midrule

\textit{Health suitability}
& MAE            & 0.0182 & \textbf{0.0128} & 0.0756 & 0.0765 \\
& RMSE           & 0.0266 & \textbf{0.0177} & 0.1052 & 0.1053 \\
& Spearman $\rho$& 0.9617 & \textbf{0.9823} & 0.1013 & 0.1182 \\

\midrule

\textit{Substitution similarity}
& MAE            & \textbf{0.0222} & 0.0344 & 0.0820 & 0.0689 \\
& RMSE           & \textbf{0.0561} & 0.0844 & 0.1720 & 0.1533 \\
& Spearman $\rho$& \textbf{0.6973} & 0.6900 & 0.6515 & 0.6699 \\

\midrule

\multicolumn{2}{l}{\textit{Weighted Multi-Objective Score}}
& \textbf{0.9628}
& 0.9164
& 0.9211
& 0.9267 \\

\bottomrule
\end{tabular}
\end{table*}
\begin{table*}[!t]
\centering
\caption{Performance comparison under the cold-start evaluation settings. CS-F = cold-start food; CS-G = cold-start household group.}
\label{tab:coldstart_results}
\begin{tabular}{llcccccccc}
\toprule
\multicolumn{2}{l}{\textbf{Metric}}
& \multicolumn{2}{c}{\textbf{HeteroGNN}}
& \multicolumn{2}{c}{\textbf{MLP}}
& \multicolumn{2}{c}{\textbf{BPR}}
& \multicolumn{2}{c}{\textbf{LightGCN}}\\

\cmidrule(lr){3-4}
\cmidrule(lr){5-6}
\cmidrule(lr){7-8}
\cmidrule(lr){9-10}

& & CS-F & CS-G & CS-F & CS-G & CS-F & CS-G & CS-F & CS-G\\

\midrule

\textit{Behavior affinity}
& ROC-AUC      & 0.8381 & 0.6070 & \textbf{0.8973} & \textbf{0.8971} & 0.4959 & 0.3883 & 0.4887 & 0.5323\\
& PR-AUC       & 0.6932 & 0.4367 & \textbf{0.7739} & \textbf{0.7618} & 0.3347 & 0.2749 & 0.3511 & 0.3910\\
& Precision@5  & 0.5143 & 0.2500 & \textbf{0.7857} & \textbf{0.8750} & 0.3333 & 0.1500 & 0.2238 & 0.3750\\
& Recall@5     & 0.0108 & 0.0008 & \textbf{0.0179} & \textbf{0.0024} & 0.0057 & 0.0006 & 0.0039 & 0.0009\\
& NDCG@5       & 0.4656 & 0.2500 & \textbf{0.8012} & \textbf{0.8453} & 0.3390 & 0.1517 & 0.2405 & 0.3750\\

\midrule

\textit{Health suitability}
& MAE          & 0.0492 & 0.0245 & \textbf{0.0140} & \textbf{0.0143} & 0.0772 & 0.0762 & 0.0771 & 0.0825\\
& RMSE         & 0.0616 & 0.0329 & \textbf{0.0200} & \textbf{0.0196} & 0.1071 & 0.1076 & 0.1071 & 0.1100\\
& Spearman $\rho$
               & 0.8609 & 0.9641 & \textbf{0.9807} & \textbf{0.9795} & 0.0140 & 0.0941 & 0.0151 & 0.1015\\

\midrule

\textit{Substitution similarity}
& MAE          & 0.1260 & \textbf{0.0333} & \textbf{0.0672} & 0.0491 & 0.1234 & 0.0818 & 0.1416 & 0.0751\\
& RMSE         & 0.2977 & \textbf{0.1130} & 0.1883 & 0.1521 & 0.2732 & 0.1971 & 0.3217 & 0.1511\\
& Spearman $\rho$
               & \textbf{0.6848} & \textbf{0.6932} & 0.6786 & 0.6493 & 0.6401 & 0.6216 & 0.6384 & 0.6691\\

\midrule

\multicolumn{2}{l}{\textit{Weighted Multi-Objective Score}}
& 0.7894 & 0.7666
& \textbf{0.9146} & \textbf{0.9082}
& 0.7054 & 0.6592
& 0.6907 & 0.7351\\

\bottomrule
\end{tabular}
\end{table*}
\begin{table*}[!t]
\centering
\caption{Robustness of HeteroGNN and MLP in incomplete information.}
\label{tab:sparsity}

\small
\setlength{\tabcolsep}{4pt}
\renewcommand{\arraystretch}{1.05}

\begin{tabular}{c cccc cccc}
\toprule
& \multicolumn{4}{c}{\textbf{Behavior Sparsity}}
& \multicolumn{4}{c}{\textbf{Nutrient Sparsity}} \\
\cmidrule(lr){2-5} \cmidrule(lr){6-9}
& \multicolumn{2}{c}{\textit{Behavior} ROC AUC $\uparrow$}
& \multicolumn{2}{c}{\textit{Health} MAE $\downarrow$}
& \multicolumn{2}{c}{\textit{Behavior} ROC AUC $\uparrow$}
& \multicolumn{2}{c}{\textit{Health} MAE $\downarrow$} \\
\cmidrule(lr){2-3} \cmidrule(lr){4-5}
\cmidrule(lr){6-7} \cmidrule(lr){8-9}
\textbf{Removed} & GNN & MLP & GNN & MLP & GNN & MLP & GNN & MLP \\
\midrule
0\%  & \textbf{0.9444} & 0.8890 & 0.0285 & \textbf{0.0262}
     & \textbf{0.9444} & 0.8890 & 0.0285 & \textbf{0.0262} \\
10\% & \textbf{0.9474} & 0.8889 & 0.0283 & \textbf{0.0263}
     & \textbf{0.9433} & 0.8895 & 0.0329 & \textbf{0.0254} \\
30\% & \textbf{0.9442} & 0.8889 & 0.0290 & \textbf{0.0265}
     & \textbf{0.9388} & 0.8890 & 0.0327 & \textbf{0.0285} \\
50\% & \textbf{0.9378} & 0.8889 & 0.0292 & \textbf{0.0265}
     & \textbf{0.9469} & 0.8876 & \textbf{0.0379} & 0.0481 \\
\bottomrule
\end{tabular}
\end{table*}
\begin{table*}[!t]
\centering
\caption{Correlations among the three objectives.}
\label{tab:objective_corr}
\begin{tabular}{@{}lcccccc@{}}
\toprule
 & \multicolumn{3}{c}{\textbf{Pearson}} & \multicolumn{3}{c}{\textbf{Spearman}} \\
\cmidrule(lr){2-4} \cmidrule(lr){5-7}
 & \textbf{Behavior} & \textbf{Health} & \textbf{Similarity} & \textbf{Behavior} & \textbf{Health} & \textbf{Similarity} \\
\midrule
Behavior affinity & 1.000  & $-$0.067 & 0.090  & 1.000  & 0.521 & 0.311 \\
Health suitability            & $-$0.067 & 1.000  & 0.070  & 0.521  & 1.000 & 0.421 \\
Substitution similarity   & 0.090  & 0.070  & 1.000  & 0.311  & 0.421 & 1.000 \\
\bottomrule
\end{tabular}
\end{table*}
\section{Experimental Setup}
\label{sec:setup}
We evaluate HeteroGNN's performance on the three substitution targets and the weighted score under standard and cold-start settings. Household-group node features include member count, household size, income-to-poverty ratio, SNAP share, poverty band, household size band, and health profile mixture. On the other hand, food features include nutrient values and food categories. Additionally, nutrient features include nutrient direction, reference amounts, and average importance weights across health profiles. Finally, category features include category code and the number of associated foods. 

We use three validation settings. The standard setting uses a random 70/10/20 partition of observed behavior edges, with the final 20 percent held out for evaluation. After that, we perform two cold-start settings. In the cold-start-food setting, similarity edges for held-out foods are also removed from the training graph. Only the node attributes remain to evaluate generalization without direct behavioral supervision. 

In the cold-start food setting, 1,774 food nodes are withheld from the behavior relation. Similarly, in the cold-start household-group setting, 8 of the 42 household groups are withheld from the behavior supervision. 
Finally, to compare the model's performance, we select baselines including Bayesian Personalized Ranking (BPR) \cite{rendle2009bpr}, a multi-Layer Perceptron (MLP), and a simplified graph network, namely LightGCN \cite{he2020lightgcn}.

For our final training architecture, we leverage a tree-structured Parzen Estimator sampler and an early-pruner to find the best hyperparameters \cite{watanabe_tpe_tutorial2023}, yielding a node embedding dimension of 64, 2 message-passing layers in the GNN, a dropout of 0.218, and a learning rate of 0.01.

\section{Results and Discussion}
\label{sec:results}

In the \textbf{standard} setting shown in Table \ref{tab:standard_results}, HeteroGNN achieves the highest difference in performance on behavior affinity, with an ROC-AUC of 0.9480 and a PR-AUC of 0.9027, outperforming LightGCN marginally and MLP substantially. This demonstrates how HeteroGNN leverages the good relational structure in behavior affinity prediction. While MLP leverages household and food attributes and captures nonlinear interactions, it lacks the capacity to propagate information through food–nutrient, food–category, and food–food relations. LightGCN performs more closely to HeteroGNN due to its graph information propagation, but its homogeneous filtering architecture does not explicitly differentiate among semantic relation types. HeteroGNN surpasses these baselines by maintaining relation-specific message passing and heterogeneous representation learning. 

In contrast to the behavior affinity, the health suitability prediction results show a different pattern. Here, MLP achieves better performance with an MAE of 0.0128 and a Spearman correlation of 0.982, outperforming HeteroGNN (0.0182, 0.962), while BPR and LightGCN perform worse than HeteroGNN. These results indicate that when explicit nutrient attributes are present, a feature-based MLP can predict health suitability with high accuracy and outperform strong learning mechanisms that capture relational structure.

For the substitution similarity objective, HeteroGNN and MLP exhibit comparable performance, with MAE values of 0.0222 and 0.0344, respectively, whereas BPR and LightGCN perform worse. The limited supervision set for similarity likely accounts for the lack of advantage observed with graph-based methods, as there is insufficient relational signal to propagate, allowing the simpler MLP to remain competitive.

Finally, the weighted multi-objective score for HeteroGNN is 0.9628, compared to 0.9164 for MLP, 0.9211 for BPR, and 0.9267 for LightGCN. MLP's weaker performance on behavior tasks, despite using the same node features, confirms that the graph structure encodes relational information not captured by node features alone. 

\subsection{Cold-Start Generalization}

Under \textbf{cold-start} setting, HeteroGNN's behavior affinity ROC-AUC decreases from 0.9480 to 0.8381 for unseen foods and further to 0.6070 for unseen household groups (shown in Table \ref{tab:coldstart_results}). This indicates that HeteroGNN's primary source of predictive power remains the observed group–food behavioral structure, suggesting that the model struggles when behavioral edges are absent.
Under the cold-start group setting, the behavior ROC-AUC falls to 0.6070, and the PR-AUC to 0.4367. However, in terms of health suitability, the MAE remains 0.0245, with a Spearman correlation of 0.9641; for substitution similarity, the MAE is 0.0333, with a Spearman correlation of 0.6932. 
To elaborate this trend, health suitability is primarily a function of food nutrient composition and the household's health profile, whereas substitution similarity is primarily a function of the relationship between the requested and candidate foods. Both can therefore be inferred from relatively stable feature information. Behavior affinity, however, depends on strong relational structure. When an entire group is unseen during behavior supervision, there is no direct relational evidence from which to infer its latent food preferences. 

On the other hand, the MLP's behavior affinity under cold start is stable and outperforms HeteroGNN in cold-start settings. This suggests that MLP relies heavily on static node attributes, which are available in both cold-start experiments.

We observe a similar trend in health suitability, with MLP outperforming HeteroGNN. This indicates that health suitability can be predicted from node features and structural similarity to known foods, without requiring direct behavioral edges.

\subsubsection{Fairness Across Demographic Segments}
\label{sec:fairness}
The aggregate cold-start household group ROC-AUC of 0.607 hides considerable variation across household segments. Table \ref{tab:fairness} breaks the same held-out predictions down by poverty band, SNAP participation, and household size. We observe that the generalization is best for households below the poverty
line, ROC-AUC 0.781, and worst for small households, ROC-AUC 0.324,
which is worse than a random ranker. Large households generalize moderately well (0.640), and single-person households sit in between (0.659). On the contrary, the SNAP status produces a smaller spread than the poverty band or household size. For implication, practitioners should treat cold-start behavior predictions for small household groups with
particular caution.

\begin{table}[!t]
\centering
\small
\caption{Behavior Affinity Performance by Household Demographic Segment (Cold-Start Group Split)}
\label{tab:fairness}
\begin{tabular}{@{}llc@{}}
\toprule
\textbf{Attribute} & \textbf{Level} & \textbf{ROC-AUC} \\
\midrule
\multirow{3}{*}{Poverty band}
& Above 2$\times$ poverty & 0.470 \\
& Below poverty           & 0.781 \\
& 1--2$\times$ poverty    & 0.667 \\
\midrule
\multirow{2}{*}{SNAP band}
& Non-SNAP & 0.590 \\
& SNAP     & 0.619 \\
\midrule
\multirow{3}{*}{Household size}
& Large  & 0.640 \\
& Single & 0.659 \\
& Small  & 0.324 \\
\bottomrule
\end{tabular}
\end{table}

\subsection{Sparsity Analysis}\label{ablation}
 We use sparsity analysis to address data shortages at the facility level. Charitable food banks often serve groups for whom only partial information is available. We perform different sparsity analyses. First, we randomly remove interaction edges (as opposed to completely removing all edges in a cold-start setting), which is defined as behavior sparsity. Second, we remove nutritional feature columns, a phenomenon known as nutrient sparsity. This tests the loss of relational and attribute information separately. Removing behavior relations affects behavior affinity prediction and impacts health scoring, whereas removing nutrient features impacts health scoring but can also affect behavioral affinity. Therefore, we report the ROC AUC for behavioral affinity and the MAE for health suitability in both analyses. 

The sparsity experiments (Table \ref{tab:sparsity}, using a different seed from the standard evaluation) provide an important explanation for why HeteroGNN remains useful despite this cold-start limitation. HeteroGNN outperforms MLP in behavior affinity prediction at all removal levels, with ROC AUC dropping only 0.7\% (0.9444 to 0.9378) from 0\% to 50\% removal, while MLP remains steady at approximately 0.889. On the other hand, MLP has a lower health MAE than GNN at all behavior removal levels.

Under nutrient sparsity, GNN is more robust, leading to a higher ROC AUC, while MLP remains stable but at a lower level. In terms of health suitability, MLP initially achieves a lower MAE, but its error increases sharply at 50\% removal. GNN’s robustness shows it leverages relational structure when features are lost.

\subsection{Multi-Objective Trade-offs}
Table~\ref{tab:objective_corr} reports the Pearson and Spearman correlation among the
three predicted objectives. Behavior affinity and health suitability are weakly negatively correlated, $-0.067$. This is consistent with the common practice that familiar, frequently consumed foods are not always the most nutrient-rich. Behavior affinity and similarity are weakly positively correlated ($0.090$), whereas health suitability and substitution similarity are weakly positively correlated ($0.070$). None of these correlations is strong enough to claim that one objective serves as a reliable proxy for another, which provides quantitative justification for keeping the three separate targets as proposed in our framework. However, a practical consideration is that
the weighted score in Eq. (\ref{eq:weighted_score}) is sensitive to how the three weights $\alpha_b$, $\alpha_h$, and $\alpha_s$ are set. Since the objectives are only weakly related, shifting weights can change which candidate ranks first. Hence, a downstream decision maker reviewing a candidate should set condition-driven weights for ranking.

\section{Conclusion and Future Work} \label{sec:conclusion}

In this study, we formulate food substitution as a multi-objective ranking problem with three objectives, providing a systematic framework for volunteers in charitable agencies. We develop HeteroGNN and a relational graph from public data sources. We demonstrate that HeteroGNN improves behavior affinity prediction beyond node features alone. We also demonstrate the robustness of HeteroGNN to behavioral and nutrient sparsity with substantial information loss. However, we also observe that cold-start conditions degrade the performance of behavior prediction. 

There are several limitations in this study that could be improved further. First, we consider behavioral proxies for supervision rather than observed targets because agency data are unavailable. Additionally, we use only 56 mapped directional labels for substitution similarity, which limits the model's predictive power. Moreover, the cold-start household evaluation comprises only 8 held-out groups, which needs a broader dataset for generalization. 

Future work should account for richer household context beyond the current demographic grouping. Additionally, future work should consider adaptive objective weighting, allowing agencies to focus on nutritional suitability, familiarity, or functional similarity according to the specific household and operational context.
	
	\bibliographystyle{IEEEtran}
	\bibliography{references}
	
\end{document}